\documentclass{article}
\usepackage[graphicx]{realboxes}  
\usepackage{soul,framed} 
\usepackage{cite}
\usepackage{multirow}
\usepackage{fontawesome}
\usepackage{booktabs}

\usepackage{mathpazo}
\usepackage{colortbl}
\usepackage{xcolor}
\graphicspath{{../pdf/}{../jpeg/}}
\DeclareGraphicsExtensions{.pdf,.jpeg,.png}
\usepackage{algorithm}
\usepackage{algpseudocode}
\usepackage{caption}
\usepackage{subcaption}
\usepackage[cmex10]{amsmath}
\usepackage[colorinlistoftodos]{todonotes}

\DeclareMathOperator*{\argmin}{arg\,min}

\usepackage{lettrine}

\usepackage{booktabs}  %
\usepackage{multirow}  %
\usepackage[left=1.3in, right=1.3in]{geometry}
\usepackage{siunitx}  %
\definecolor{mygray}{gray}{0.9}
\usepackage{cellspace}

\usepackage{xcolor} 
\title{Fed-Safe: Securing Federated Learning in Healthcare Against Adversarial Attacks}
\author{Erfan~Darzi$^{1,*}$, Nanna M. Sijtsema$^{2}$, P.M.A van Ooijen$^{2}$ \\
\small{$^{1}$ Harvard Medical School, Harvard University, Boston, MA, USA}\\
\small{$^{2}$Department of Radiation Oncology, University Medical Center Groningen,} \\
\small{University of Groningen, The Netherlands}\\
}
\date{} 
\begin{document}
\maketitle
\footnotetext{*Corresponding author: erfandarzi@gmail.com}
\begin{abstract}

This paper explores the security aspects of federated learning applications in medical image analysis. Current robustness-oriented methods like adversarial training, secure aggregation, and homomorphic encryption often risk privacy compromises. The central aim is to defend the network against potential privacy breaches while maintaining model robustness against adversarial manipulations. We show that incorporating distributed noise, grounded in the privacy guarantees in federated settings, enables the development of a adversarially robust model that also meets federated privacy standards. We conducted comprehensive evaluations across diverse attack scenarios, parameters, and use cases in cancer imaging, concentrating on pathology, meningioma, and glioma. The results reveal that the incorporation of distributed noise allows for the attainment of security levels comparable to those of conventional adversarial training while requiring fewer retraining samples to establish a robust model.

\end{abstract}
\section{Introduction}
\lettrine{F}{}ederated learning is a popular method for medical image analysis, due to its promise of high accuracy while maintaining data privacy. One of its key advantages over central storage of data is that it solves problems related to data governance and privacy by training algorithms without exchanging sensitive information. However, despite its limited communication overhead compared to central storage methods, federated learning is not without its privacy and security challenges. For example, malicious clients might be able to acquire sensitive information from other clients or disrupt the training process. These challenges hinder the large-scale deployment of federated learning networks in medical image applications.

This growing threat from potential adversaries has incentivized research on enhancing the robustness of federated networks. Various security-enhancing methods have emerged, including adversarial training, secure aggregation, and homomorphic encryption. Among these, adversarial training is the most popular technique for mitigating the impact of adversarial attacks. However, these approaches come with a fundamental drawback: they compromise the privacy of both the model and the data~\cite{zhang2022privacy, mejia2019robust, song2019privacy}. Studies have shown that adversarial training can amplify the efficacy of some privacy attacks which are otherwise deemed ineffective against traditionally trained models. For example, one study reported that adversarial training made it significantly easier to perform multiple types of attacks, such as model inversion, data reconstruction, and GAN-based attacks. The effectiveness of these attacks increased by an order of magnitude when models were adversarially trained~\cite{mejia2019robust}.

This compromise between security and privacy is commonly known as the "robustness-privacy tradeoff of adversarial training." It raises an important question: Is the protection against adversarial attacks worth the high cost to privacy?

\subsection*{Our Contributions}
In this paper, we provide certified and well-understood privacy guarantees to adapt adversarial training in a federated environment. Our approach aims to maintain adversarial robustness while minimizing privacy risks. Central to our method is the leverage of inherent privacy guarantees in federated settings which enables us attain privacy-preserving  adversarial training.

\textbf{Adversarial robustness 
}We demonstrate that our federated adversarial adaptation method achieves a level of robustness comparable to traditional adversarial training. Specifically, we find that we can match or even surpass the robustness offered by state-of-the-art adversarial training in various scenarios.

\textbf{Guaranteed Privacy Through Distributed Noise}
We demonstrate that our approach not only enhances adversarial robustness but also maintains privacy. Our modified adversarial training method utilizes distributed noise to ensure certified privacy guarantees.

\textbf{Efficient Data Utilization
}Our method is designed to minimize data requirements for model retraining, which is important in medical data where data sharing is a burden. Compared to traditional adversarial training approaches, our adversarial adaptation strategy requires significantly less data to train the global model, thereby achieving desired levels of robustness with fewer data samples.


\section{Related Works}
While numerous studies have been conducted on the topic of federated learning within the realm of medical image analysis, they have largely ignored the aspect of security and privacy of these setups. Similarly, though there are some research on adversarial robustness focused on medical images, these primarily examine scenarios within a centralized computational framework. As far as we are aware, our research stands as an initial effort in combining these two important areas—federated learning and robustness—in the specific context of medical imaging.

\textbf{Federated Learning for Medical Image Analysis.} Federated learning has been  explored in multiple studies for medical image analysis. For instance, Liu et al. proposed a federated learning system for medical image classification that achieved an accuracy of 89.17\% for chest X-Ray,\cite{bouacida2021vulnerabilities} using multiple hospitals as clients\cite{liu2020experiments}. Similarly, Malik et al. proposed a secure federated learning framework for medical image classification, using a secure multi-party computation protocol for secure aggregation and achieving an accuracy of 98.45\% on the Chest X-Ray dataset\cite{malik2023dmfl_net}.

\textbf{Adversarial Attacks on Medical Images.}  Studies have shown that medical images are vulnerable to adversarial attacks in a centralized setting. For instance, Han et al. \cite{han2020deep} proposed a deep learning-based system for classifying cardiac diseases from ECG signals and assessed its vulnerability to various white-box and black-box adversarial attacks. Similarly, Rodriguez et al.'s proposed system for classifying chest x-ray abnormalities \cite{rodriguez2022role} was found to be highly vulnerable to different types of adversarial attacks. Bortsova et al. further studied the adversarial attacks and attack detection methods, in medical images within a centralized setting \cite{bortsova2021adversarial}.

\textbf{Secure Federated Learning}  (SFL) has been widely studied, with research focusing on both privacy and robustness. For example, So et al. proposed a secure aggregation protocol for SFL to protect against various attack scenarios \cite{so2022lightsecagg}. Chen et al. presented an end-to-end secure aggregation protocol to maintain client privacy while preserving model accuracy \cite{chen2022poisson}. Studies concerning encryption and privacy-preserving approaches, such as differential privacy, have also been conducted. Ma et al., for instance, proposed a front-end framework using multi-key encryption \cite{ma2022privacy}, while Zhang et al. created a platform using homomorphic encryption for secure aggregation without considering DP or encrypted inference \cite{zhang2022homomorphic}. Sheller et al. implemented a privacy-preserving FL segmentation model without using DP or assessing its robustness\cite{sheller2020federated}, Vithana et al. showcased a segmentation task but relied on a different technique (sparse vector) instead of DP \cite{vithana2022model}, and Adnan et al. provided a framework for FL with differential privacy concerning pathology slides without covering secure aggregation or encrypted inference \cite{adnan2022federated}. Adversarial training has been applied as an optimization technique for individual clients in order to secure federated networks; however, this also increases computational burden for all participants, potentially leading to lack of convergence due to heterogeneity among models \cite{bagdasaryan2020backdoor}.
\section{Adversarial adaptation with distributed noise }
In federated learning, clients and a central server communicate via uplink and downlink channels. The uplink channel allows clients to upload local model parameters to the server, with the model designed to enforce \((\epsilon,\delta)\)-Differential Privacy (DP).

Parameters or functions of these parameters are uploaded, reducing exposure of sensitive data. Differential privacy metrics evaluate the privacy cost of this data transfer. Gaussian noise, scaled by query sensitivity, preserves the privacy of uploaded models. Conversely, the downlink channel disseminates the global model to the clients. Though this is an aggregate, it remains susceptible to reverse-engineering attacks that could disclose sensitive client data. The global privacy preserving requirements for federated learning has been analyzed by Wei et al\cite{wei2020federated}. For the uplink, a clipping method bounds each client's training parameters \(\mathbf{w}_{i}\) with a value \(C\). Local training at client \(i\) is defined as:
\begin{equation}
\label{localtrain}
s_{\text{U}}^{\mathcal D_i} := \mathbf{w}_{i}=\arg\min_{\mathbf{w}}{F_{i}(\mathbf{w}, \mathcal D_i)} = \frac{1}{\vert \mathcal D_i \vert}\sum_{j = 1}^{\vert \mathcal D_i \vert}\arg\min_{\mathbf{w}}{F_{i}(\mathbf{w}, \mathcal D_{i,j})},
\end{equation}
Sensitivity \(\Delta s_{\text{U}}^{\mathcal D_i}\) is computed as:
\begin{equation}
\label{SensitivityforUP}
\Delta s_{\text{U}}^{\mathcal D_i} = \max_{\mathcal D_i, \mathcal D_i'}{\Vert s_{\text{U}}^{\mathcal D_i} - s_{\text{U}}^{\mathcal D_i'} \Vert} = \frac{2C}{\vert \mathcal D_i \vert},
\end{equation}
Global sensitivity in the uplink is denoted as \(\Delta s_{\text{up}}\):
\begin{equation}
\Delta s_{\text{up}} = \max_{i}\{\Delta s_{\text{up}}^{\mathcal{D}_i}\}, \quad \forall i.
\end{equation}
To control this sensitivity, each client should have a local dataset of at least size \(m\), leading to \(\Delta s_{\text{up}} = \frac{2C}{m}\).

Noise scale \(\sigma_{\text{up}}\) is introduced, and adjusted for multiple \(L\) exposures as \(\sigma_{\text{up}} = \frac{cL\Delta s_{\text{up}}}{\epsilon}\).

In the downlink channel, the aggregated dataset \(\mathcal{D}_i\) is:
\begin{equation}
\label{DLfunction}
s_{\text{down}}^{\mathcal D_i} \equiv \mathbf{w} = p_1\mathbf{w}_1 + \cdots + p_i\mathbf{w}_i + \cdots + p_N\mathbf{w}_N,
\end{equation}
where \(1 \leq i \leq N\) and \(\mathbf{w}\) represents the unified parameters that are subsequently disseminated by the server to the clients \cite{wei2020federated}.
\\\textbf{Adversarial Robustness} In the context of federated learning, adversarial robustness is crucial for enhancing the resilience of the federated model against adversarial attacks. These attacks can compromise the privacy and performance of the global model. To address this challenge, adversarial training is often employed. This involves altering the training data with small, malicious perturbations to deceive the model, aiming to make the model more resilient in adversarial conditions.

A common measure for robustness is the adversarial loss \( \mathcal{L}_{\text{adv}} \), defined as:

\begin{equation}
\label{adversarialLoss}
\mathcal{L}_{\text{adv}}(\mathbf{w}, \mathcal{D}) = \max_{\delta \in \Delta} \mathcal{L} (\mathbf{w}, \mathcal{D} + \delta),
\end{equation}

Here, \( \Delta \) is the set of allowable perturbations to the dataset \( \mathcal{D} \), and \( \mathcal{L} \) is the original loss function.  During adversarial training, the objective is to minimize this adversarial loss by iteratively updating the model parameters \( \mathbf{w} \). Adversarial training could be either done in local training, or in global server level with a separate server dataset . This process enhances the robustness of the federated learning model against adversarial attacks.

\subsection{Distributed noise}
The challenge of preserving privacy while training local models is a significant concern, especially given that these models are often vulnerable to reverse engineering and subsequent privacy breaches. Typically, these models are trained without explicit privacy safeguards, thereby exposing them to potential adversarial attacks aimed at decoding data from individual clients. To mitigate these risks, we introduce a dual-faceted approach that enhances both security and privacy.

Our distributed noise model serves as a robust solution to this challenge. It comprises two key components:

First, we generate a series of synthetic samples to perform adversarial adaptation on the global server. Additionally, we add distributed Gaussian noise to the downlink channels to the client. By minimizing the adversarial adaptation loss function, the network becomes resilient to potential adversaries that may be present in the clients. Furthermore, this method of noise-based protection does not interfere with local training. The adversarial adaptation of the global model facilitates the learning of the data distribution across all client models, thereby increasing its resilience against potential threats.

\textbf{Model Aggregation}
In federated learning networks, a central server aggregates updates from multiple clients. For this study, we employ Federated Averaging as our aggregation model, although the defense technique discussed is applicable to other aggregation models as well. To further bolster the privacy of our model updates, we introduce distributed noisy communication channels.

\begin{algorithm}
\caption{Federated adversarial adaptation}
\label{algo1}
\begin{algorithmic}[1]
    \Require $X \in \mathcal{X}$, $Y \in \{-1, 1\}$, $g_{\beta}$: $\mathcal{X} \to \{-1, 1\}$, $\epsilon$: privacy parameter, $\eta$: bound, $\alpha$: step size
    \State Initialize parameters: $\beta = \beta_0$ and $T = 0$
    \While{$T < T_{max}$}
        \For{each client}
            \State Perform local federated training with differential privacy 
            \State $g_{\beta} \leftarrow g_{\beta_T}$
        \EndFor
        \State Aggregate the models from all clients to create the global model
        \State Update parameters: $\beta = \beta_{T+1}$ and $T = T+1$
    \EndWhile
    \State Generate adversarial samples using the global model:
    \begin{equation*}
    x_{adv} = x + \epsilon\ \text{sign}(\nabla_x\mathcal{L}(\theta,x,y))
    \end{equation*}
    \State Perform federated adversarial adaptation $g_{\beta_{T+1}}$ with data set $X_T$, $Y_T$ by performing the following optimization: 
    \begin{equation*}
    g_{\beta_i} = \argmin_{g_{\beta}} \mathcal{L}(g_{\beta}, D_i \cup D_i^{adv}) 
    \end{equation*}
    \For{$k \in \{1,\dots,N\}$}
        \State Add distributed noise $\mathcal{N}(0, \sigma^2)$ to $g_{\beta_{T+1}}$
        \State Receive the noisy model $\hat{g}_{\beta_{T+1}}^{(k)}$
    \EndFor
\end{algorithmic}
\end{algorithm}

\textbf{Generating Adversarial Samples}  Within the client-side architecture, the global model obtained from the central server undergoes an update process based on a local dataset. To achieve this, an optimization algorithm is applied to the model. Adversarial samples are generated using two distinct methods: First, the global model is tested against a specific dataset to produce these samples. The samples are generated using two methods, FGSM and PGD.
The \textit{Fast Gradient Sign Method} (FGSM) is a technique aimed at generating adversarial sample. It does so by taking a specific input \(x \in \mathcal{X}\), where \(\mathcal{X}\) is the domain of all possible samples, and introducing controlled perturbations to it. The key to this method is the use of the gradient \(\nabla\) of the model's loss function \(\mathcal{L}\)\cite{kurakin2018adversarial}. The perturbed or adversarial sample \(x_{\text{adv}}\) is created in a single step as follows:
\begin{equation} \label{eq:FGSM}
x_{\text{adv}} = x + \epsilon\ \text{sign}(\nabla_x\mathcal{L}(\theta,x,y)),
\end{equation}
Here, \(\epsilon\) is a small positive constant that adjusts the extent of the perturbation. The variable \(\theta\) represents the trainable parameters of the machine learning model. This approach is computationally efficient but may not produce the most potent adversarial sample.

In contrast to FGSM, the \textit{Projected Gradient Descent Attack} (PGD) provides a more robust method for creating adversarial sample. It builds upon the foundational principles of FGSM but employs a multi-step iterative process\cite{madry2017towards,liu2022threats}. The mathematical formulation for generating an adversarial sample using PGD is:
\begin{equation} \label{eq:PGD}
x^{i+1} = \text{clip}(x^i + \alpha\ \text{sign}(\nabla_{x^i}\mathcal{L}(\theta,x^i,y)),
\end{equation}
In this equation, \(i\) indicates the current iteration, and \(\alpha\) is a tunable constant that affects the magnitude of the disturbance at each iteration. The function \text{clip}(\(\cdot\)) ensures that the perturbations stay within a certain limit, usually defined by \(|\alpha\ \text{sign}(\nabla_{x^i}\mathcal{L}(\theta,x^i,y))| \leq \eta\), where \(\eta\) is another positive constant. PGD tends to generate stronger adversarial samples at the cost of being computationally more demanding\cite{liu2022threats}.

Suppose every instance $x \sim D$ corresponds to a fixed label $y \in \{-1, +1\}$ and the label of $+1$ indicates that instance $x$ was generated by an adversary. To evade the classifier, adversary would generate an alternative instance $x' \in X$, for which $g(x') = -1$. 
Traditionally in the field of machine learning, the focus is often on minimizing the empirical risk, which can be formally expressed as:

\begin{equation}
\label{E:modified_loss_function}
\argmin_{\theta} \mathcal{L}(\theta) = \sum_{i} \ell(f_{\theta}(x_i), y_i)
\end{equation}

In this equation, $\ell(\hat{y}, y)$ represents the loss incurred when the predicted outcome is $\hat{y}$ while the actual label is $y$. Here, $x_i$ refers to the input feature vector and $y_i$ corresponds to the associated label within the training dataset. 

However, the standard approach may not suffice in adversarial settings. When dealing with malicious entities, the aim isn't just to classify instances correctly but also to anticipate evasive actions by these entities. Specifically, a sample classified as malicious, denoted by $f_{\beta}(x) = +1$ in our notation, might be altered by an adversary to bypass detection.

To capture this adversarial behavior, consider a function defined as $x^{adv}(\beta, x)$. Given a parameter vector $\beta$ and an original feature vector $x$, this function returns a modified feature vector $x'$ that represents the instance after adversarial modification.

\textbf{Federated Adversarial Adaptation} After the adversarial samples are generated,
they are used to perform federated adversarial adaptation. The adaptation process follows optimizing of an augmented dataset, to train a robust model against 
adversarial attacks. The new model parameters are then sent to all the clients for evaluation. Given the adversarial nature of the scenario, wherein the adversary alters malicious instances based on the functions's input, the defender's effective risk can no longer be represented solely by Equation~\ref{E:modified_loss_function}. Instead, we need to factor in the potential adversarial responses. As a result, the global model now aims to reduce the so-called federated adversarial adaptation loss. This loss, specific to the training data, is expressed as:

\begin{equation}
\label{E:advloss}
\min_\theta \mathcal{L}_A(\theta; x^{adv}_{i}) = \sum_{i:y_i = -1} l(g_\theta(x_i),-1) + \sum_{i:y_i = +1} l(g_\theta(x^{adv}_{i}(\theta,x_i),+1) 
\end{equation}

In this equation, $g_\theta(x)$ symbolizes the output of the classifier, while $\mathcal{L}_A(\theta; x^{adv})$ represents the adversarial adaptation loss. The function \( x^{adv}(\theta,x) \) yields an adversarially modified feature vector for the given input \( x \) based on parameter \( \theta \).

The risk function detailed in Equation~\ref{E:advloss} is constructed in a way that it remains versatile. It can adapt to a variety of scenarios, given that we aren't making specific assumptions about the attacker's function, denoted by ${x^{adv}}$.

\textbf{Empirical risk evaluation} To understand the outcome when the algorithm comes to an end, we first identify the empirical risk during the final iteration of the federated adversarial adaptation:

\begin{equation}
\label{E:radloss}
\mathcal{L}^R_N(\beta, {x^{adv}}) = \sum_{i \in D \cup N} l(g_\beta(x_i),y_i) 
\end{equation}
Here, $N$ is a collective set of data points, integrated by the algorithm.
With this, we can now discern the relationship between \(\mathcal{L}^R_N(\beta, {x^{adv}})\) and the optimal adversarial adaptation loss \(\mathcal{L}_A^*({x^{adv}}) = \min_\beta \mathcal{L}_A(\beta, {x^{adv}})\). Mathematically, for any parameter set \(\beta\):

\begin{align*}
\mathcal{L}^R_N(\beta, {x^{adv}}) &\ge \mathcal{L}^R_N(\bar{\beta}, {x^{adv}}) \\
&= \sum_{i: y_i = -1} l(g_{\bar{\beta}}(x_i),-1) +\sum_{i:y_i = +1} \sum_{j \in N_i \cup x_i} l(g_{\bar{\beta}}(x_i),+1) \\
&\ge \sum_{i: y_i = -1} l(g_{\bar{\beta}}(x_i),-1) +\sum_{i:y_i = +1} l(g_{\bar{\beta}}({x^{adv}}(\bar{\beta},x_i)),+1) \\
&\ge \min_\beta \mathcal{L}_A(\beta; {x^{adv}}) = \mathcal{L}_A^*({x^{adv}})
\end{align*}

The logic behind the second inequality stems from the last iteration of our algorithm, during which no new values are added. This implies that the adversarial samples \({x^{adv}}(\beta,x_i)\) fall within \(N_i\) for every \(i\) belonging to \(I_{adv}\).
 This demonstrates how retraining the global model with adversarial samples can improve the robustness of the model against adversarial attacks. The proof shows that the  empirical risk in the last iteration of the algorithm is lower than the adversarial risk, and thus the model is more robust against adversarial attacks. 

\include{proposed_method.tex}
\section{Experiments and Results}

\subsection{Datasets}
To simulate real-world conditions, we used non-independent and non-identically distributed (non-IID) datasets across multiple clients. Data were divided into non-overlapping chunks, each representing a segment of the general distribution, to ensure that each client's dataset is distinct, fulfilling the requirements of federated learning scenarios.

In the domain of brain cancer classification, we utilized a public dataset of MRI scans available on Kaggle \cite{sartaj}. This dataset comprises four classes: three types of brain tumors and one healthy class. We used 2870 training images and 394 testing images, with 1437 and 1426 images specifically for Meningioma and Glioma detection, respectively. The images were resized to 100$\times$100 pixels and underwent rotation, flipping, and normalization for data augmentation.

For histopathologic cancer detection, we sourced high-resolution pathology slices from previous work \cite{veeling2018rotation}. The dataset consists of 2150 images with a 62\%/38\% training/testing split. Metastatic tissue, when present, is centered within a $32\times32$ pixel region. These images were also transformed via horizontal flipping and normalization to enhance the dataset.
\begin{figure}[t]
    \centering
    \includegraphics[width=1\textwidth]{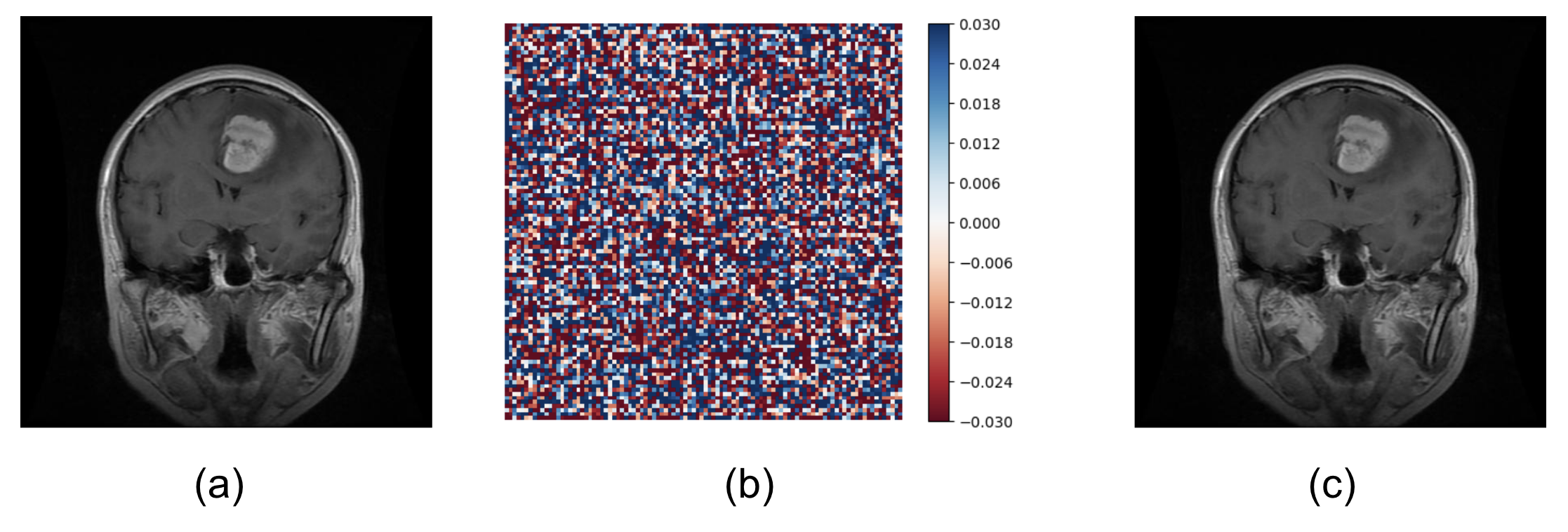}
    \caption{(a) Original unmanipulated meningioma MRI. (b) Adversarial noise generated with FGSM method with $\epsilon$ = 0.03 . (c) Manipulated meningioma MRI (original image + noise), i.e., adversarial example.}
    \label{fig:adversarial_examples}
\end{figure}

\subsection{Experimental Setup }
\textbf{Federated setup}:
A Convolutional Neural Network (CNN) with six layers of convolution stacks, five fully connected layers and ReLU activation function is used as the deep learning model (with 0.25 dropout parameter). The model is trained with Cross-Entropy loss via SGD optimizer. For federated setting, three clients equipped with random data with non-IID data distribution. Federated averaging (FedAVG) method is used to aggregate the models with 20 epochs per communication round and 50 federated rounds in total, weighed based on the size of the training dataset.
\\\textbf{Evaluation Metrics }:
We use two metrics to evaluate the performance of adversarial attacks, i.e. clean accuracy and attack success rate. Clean accuracy is defined as the performance of models on uncorrupted test images. Attack success rate (ASR) measures how much an adversary can change the predicted labels produced by each model. Formally, ASR is defined as:
\begin{equation}
    ASR = \frac{1}{N}\sum\limits_{i=1}^{N} (Pre\text{-}attack\;label_{i}\neq Post\text{-}attack\;label_{i})
\end{equation}
where $N$ is the number of images in the attack set. Attack accuracy (AA) shows the accuracy of an adversary after performing an attack, and transferability measures the effectiveness of attacks on target models.  Figure \ref{fig:adversarial_examples} presents an illustration of a slice of a Meningioma MRI scan that has been  adversarially manipulated.
\subsection{Robustness with federated adversarial adaptation}

\begin{figure*}[b!]
    \centering
     \begin{subfigure}[b]{1.5\textwidth}
         \centering
        \hspace{-170pt} \includegraphics[width=0.49\textwidth]{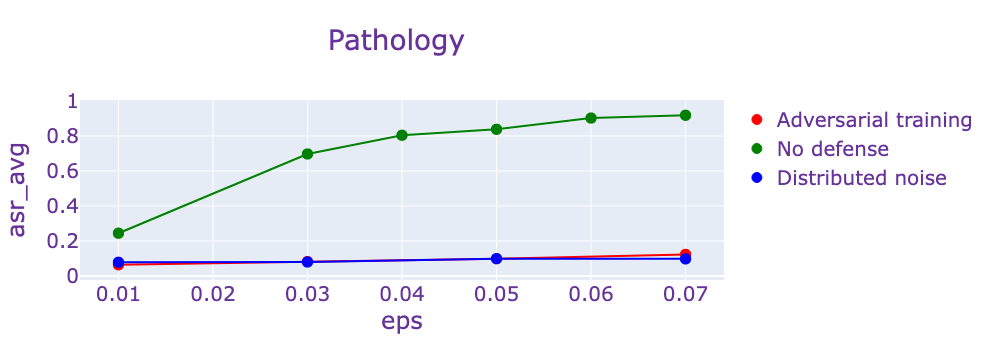}        \hspace{-85pt}       \includegraphics[width=0.49\textwidth]{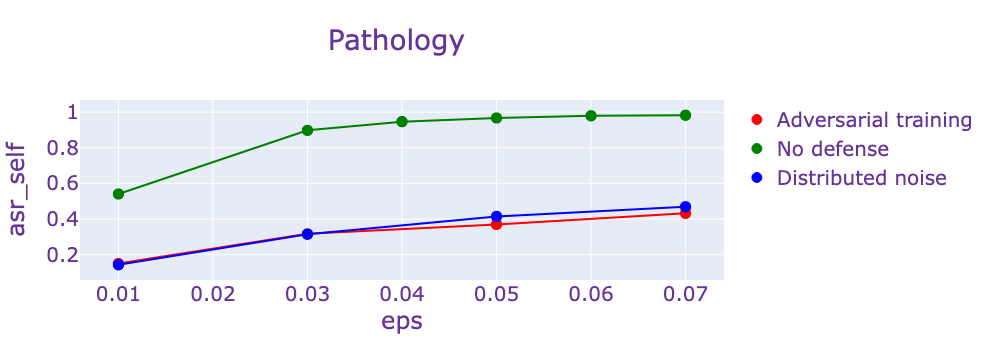}      \centering
         \label{fig:AASR benign-EPS}
     \end{subfigure}
        \begin{subfigure}[b] {1.5\textwidth}
         \centering
        \hspace{-170pt} \includegraphics[width=0.49\textwidth]{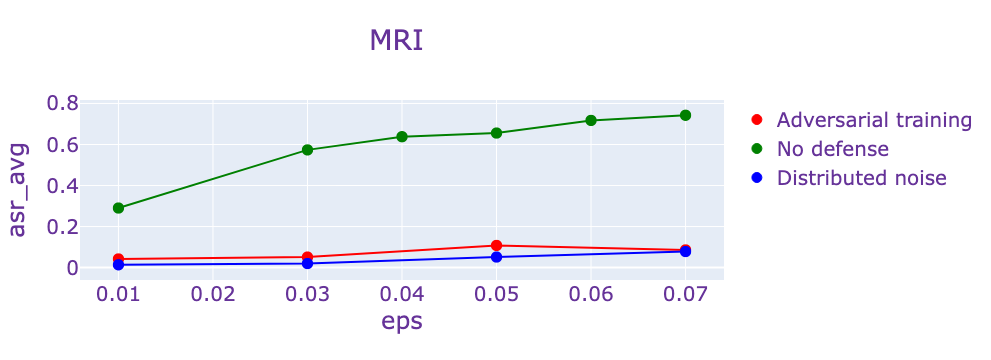}        \hspace{-85pt}       \includegraphics[width=0.49\textwidth]{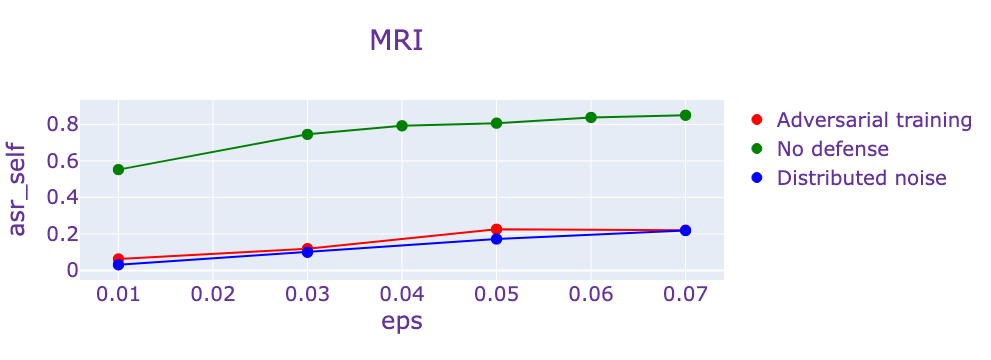}              
         \label{fig:AASR benign-EPS}
     \end{subfigure}
     
      \centering
     \begin{subfigure}[b]{1.5\textwidth}
         \centering
        \hspace{-170pt} \includegraphics[width=0.49\textwidth]{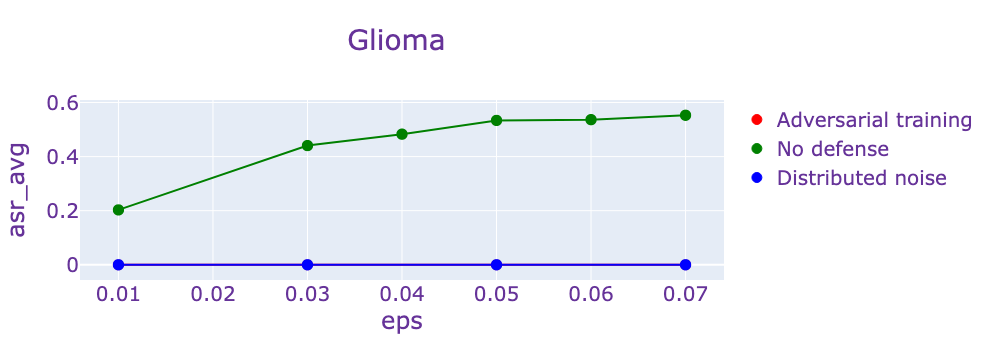}        \hspace{-85pt}       \includegraphics[width=0.49\textwidth]{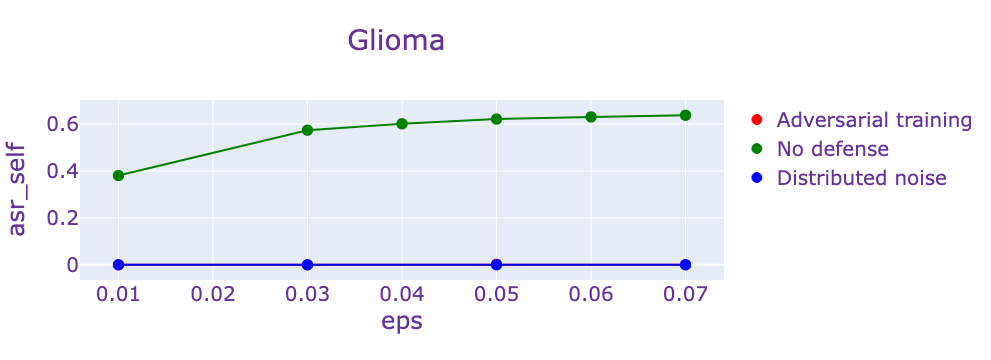}      \centering
         \label{fig:AASR benign-EPS}
     \end{subfigure}
     \hfill
        \caption{Effect of Error perturbation degree $\epsilon$ on attack transferability.PGD attack is performed.ASR is calculated on benign (left column) and the adversary client (right column). The higher ASR on benign clients shows higher transferability}
        \label{fig: epsilon graphs}
\end{figure*}

In this section, we conduct a comparative analysis between our proposed method and traditional adversarial training. The evaluation aims to illustrate the relative efficiencies of the methods in various scenarios.
Figure \ref{fig: epsilon graphs} and Table \ref{tab:my_label_fancy} depict the comparison of distributed noise and adversarial training across three classification tasks, under different attack settings, characterized by varying degrees and steps of perturbation. Distributed noise training was able to reduce the attacker's self-ASR by up to 8\%, and the attacker's average ASR by up to 27\%, compared to the  adversarial training method.

\begin{table}[h]

    \centering
    \caption{Average values of ASR(self/avg) for different datasets as a result of distributed noise training and adversarial training technique. Lower value of ASR indicates a better defense against adversarial attacks.}
    \label{tab:my_label_fancy}
    \begin{tabular}{lccc}
    \toprule
    \multicolumn{1}{c}{} & \multicolumn{3}{c}{\textbf{Training Method}} \\
    \cmidrule(l{4pt}){2-4}
    \textbf{Dataset} & \textbf{Step} & \textbf{Adversarial training} & \textbf{Distributed Noise} \\
    && \textbf{(ASR Self/Avg)} & \textbf{(ASR Self/Avg)} \\
    \midrule
    \textbf{Glioma} & 0.005 & \cellcolor[gray]{0.9}0.03/0.00 & \cellcolor[gray]{0.9}0.03/0.00 \\
    & 0.012 & 0.00/0.00 & 0.00/0.00 \\
    & 0.017 & \cellcolor[gray]{0.9}0.00/0.00 & \cellcolor[gray]{0.9}0.00/0.00 \\
    & 0.05 & 0.03/0.00 & 0.00/0.00 \\
    \midrule
    \textbf{MRI} & 0.005 & \cellcolor[gray]{0.9}0.12/0.05 & \cellcolor[gray]{0.9}0.10/0.02 \\
    & 0.012 & 0.22/0.09 & 0.22/0.08 \\
    & 0.017 & \cellcolor[gray]{0.9}0.25/0.11 & \cellcolor[gray]{0.9}0.23/0.04 \\
    & 0.05 & 0.46/0.27 & 0.41/0.15 \\
    \midrule
    \textbf{Pathology} & 0.005 & \cellcolor[gray]{0.9}0.32/0.08 & \cellcolor[gray]{0.9}0.31/0.08 \\
    & 0.012 & 0.43/0.12 & 0.47/0.10 \\
    & 0.017 & \cellcolor[gray]{0.9}0.46/0.17 & \cellcolor[gray]{0.9}0.49/0.12 \\
    & 0.05 & 0.54/0.32 & 0.65/0.34 \\
    \bottomrule
    \end{tabular}
\end{table}

The results also indicated that higher values of the perturbation parameter led to a higher attack success rate in all scenarios. In both defense models, the self ASR was around two to four times higher than the average ASR, indicating that the differences between the models in a federated learning network can lead to significant differences in the success of an attack. In contrast, the error rate of the undefended model was similar for the adversarial client and other clients, indicating high transferability of the attack.\\

\begin{table}[h]
    \centering
    \caption{Effect of DP on the performance of distributed noise and adversarial training. The average attack success rate (ASR) values are shown for each dataset.}
    \label{tab:perturb_param}
    \begin{tabular}{lccc}
    \toprule
    \multicolumn{1}{c}{} & \multicolumn{3}{c}{\textbf{No DP, DP, and $\epsilon$}} \\
    \cmidrule(l{4pt}){2-4}
    \multicolumn{1}{c}{} & No DP & DP & $\epsilon$ \\
    \midrule
    Pathology & \cellcolor[gray]{0.9} 10.21\% & \cellcolor[gray]{0.9} 5.09\% & \cellcolor[gray]{0.9} 0.01 \\
    & 9.97\% & 8.72\% & 0.03 \\
    \cmidrule(l{4pt}){2-4}
    & \cellcolor[gray]{0.9} 9.53\% & \cellcolor[gray]{0.9} 7.87\% & \cellcolor[gray]{0.9} 0.05 \\
    & 19.04\% & 16.67\% & 0.07 \\
    \midrule
    \textbf{Average} & \cellcolor[gray]{0.9} \textbf{12.19\%} & \cellcolor[gray]{0.9} \textbf{9.58\%} & \cellcolor[gray]{0.9} \\
    \midrule
    MRI & \cellcolor[gray]{0.9}2.74\% & \cellcolor[gray]{0.9}0.31\% & \cellcolor[gray]{0.9} 0.01 \\
    & 3.92\% & 0.32\% & 0.03 \\
    \cmidrule(l{4pt}){2-4}
    & \cellcolor[gray]{0.9} 12.67\% & \cellcolor[gray]{0.9} 0.32\% & \cellcolor[gray]{0.9} 0.05 \\
    & 6.29\% & 0.32\% & 0.07 \\
    \midrule
    \textbf{Average} & \cellcolor[gray]{0.9} \textbf{6.41\%} & \cellcolor[gray]{0.9} \textbf{0.32\%} & \cellcolor[gray]{0.9} \\
    \bottomrule
    \end{tabular}
\end{table}

\begin{figure}[]
    \centering
     \begin{subfigure}[b]{1.7\textwidth}
         \centering
         \hspace{-260pt}  
         \includegraphics[width=0.32\textwidth]{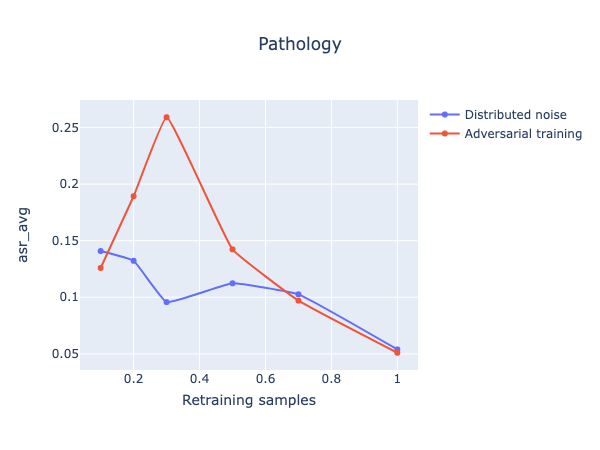}        \hspace{-77pt}       \includegraphics[width=0.32\textwidth]{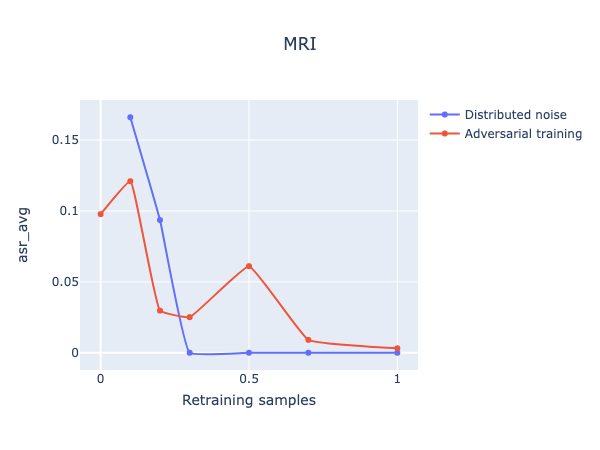} 
         \hspace{-77pt} 
          \includegraphics[width=0.32\textwidth]{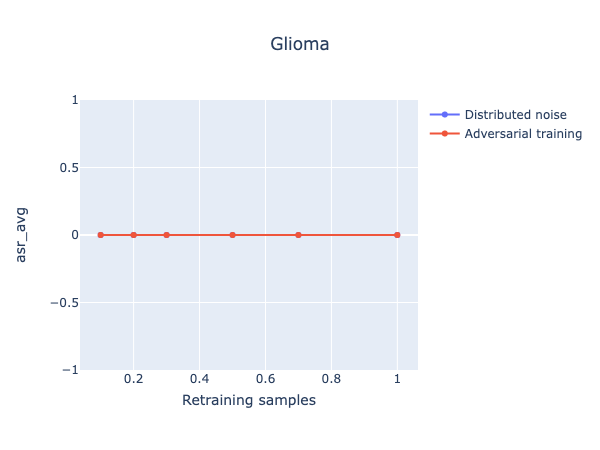} 
          \centering
     
         \label{fig:AASR benign-EPS}
     \end{subfigure}
       \centering
     \begin{subfigure}[b]{1.7\textwidth}
         \centering

         \hspace{-260pt}  
    \includegraphics[width=0.32\textwidth]{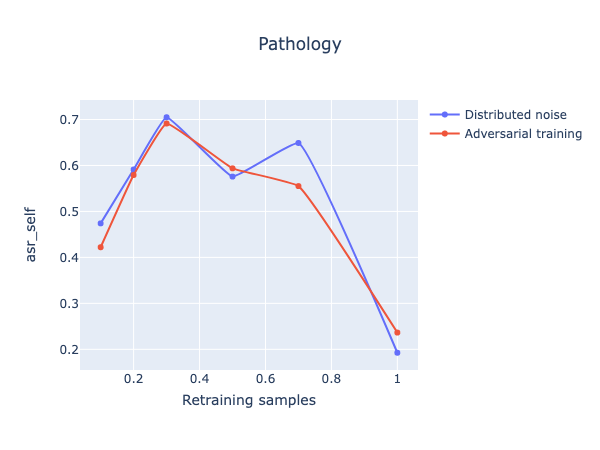}        \hspace{-77pt}       \includegraphics[width=0.32\textwidth]{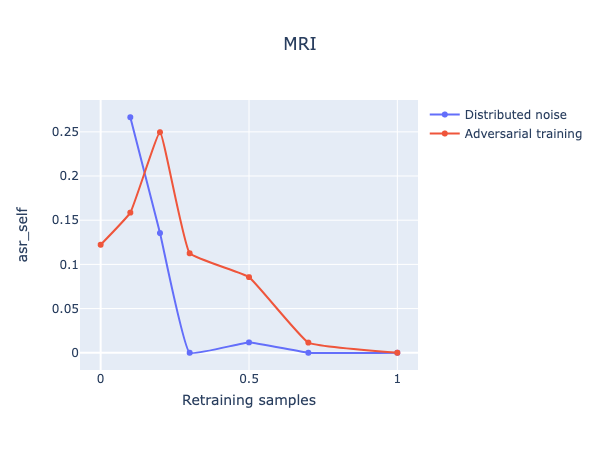}            
         \hspace{-77pt}    
\includegraphics[width=0.32\textwidth]{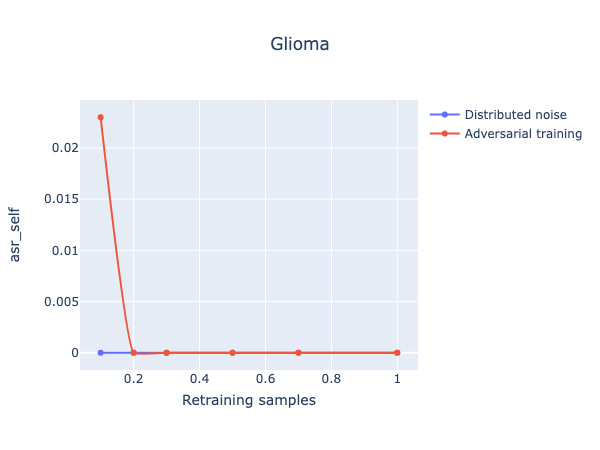} 
         \label{fig:AASR benign-EPS}
     \end{subfigure}
        \caption{Effect of the number of adversarial samples on the average attack success rate, for benign and adversarial clients. The results show that using more retraining samples in the global model decreased the attack success rate in all scenarios,  with distributed noise outperforming adversarial training.}
        \label{fig: retraining samples }
\end{figure}

\textbf{Distributed noise requires fewer training samples} Figure~\ref{fig: retraining samples } shows the effect of the number of adversarial adaptation samples on the performance of distributed noise and adversarial training, especially in the ASR for the benign clients. We evaluated from 10,20,30,50,70 and 100 percent of the samples, and the results indicate that using more retraining samples in the global model leads to a lower attack success rate in all scenarios. Using all of the training samples led to almost perfect defense in all scenarios, while distributed noise outperformed adversarial training and achieved perfect defense with much fewer samples. For example, using only 20\% of the training samples would result can reduce the attackers success on benign clients to 12.2\% pathology, 9.7\% in Meningioma, and 0.0\% in Glioma usecase.

\textbf{Effect of differential privacy on robustness} While methods like adversarial training have a compromise on privacy, our findings suggest that DP can actually enhance robustness. Table \ref{tab:perturb_param} shows the results of the experiments conducted to evaluate the effect of DP on the defense methods. In both defense methods, adversarial training and distributed noise, DP reduced the average attack success rate (ASR) and hence improved the defense. For all attack values, the Glioma dataset resulted in perfect defense with ASR=0.0, and DP enabled defense for MRI was also almost perfect with both defense methods, resulting in average ASR= 0.32 percent. For the Pathology dataset, the average ASR was reduced from 12.19 percent to 9.58 percent when differential privacy was used.   
Our experiments showed that using distributed noise training can reduce the attacker's self-ASR by up to 8\% and the attacker's average ASR by up to 27\%, across three different datasets. Additionally, we observed that the effectiveness of distributed noise is dependent on the step size used for the attack, with smaller steps leading to greater effectiveness. 

We found that using more retraining samples in the global model can lead to a lower attack success rate, and that the addition of differential privacy can improve defense performance. These findings are consistent with the existing literature on adversarial defense, emphasizing the role of retraining samples and differential privacy as viable defense mechanisms. However, the inherent trade-off between differential privacy and clean accuracy necessitates further exploration. A comparative analysis delineating the effectiveness, computational cost, and implementation feasibility of our technique against other defense mechanisms remains a subject for future research.
\newpage
\section{Conclusion }
This paper focuses on robustness of federated learning networks against adversarial attacks through the implementation of distributed noise. Our methodology offers defense against a diverse array of adversarial scenarios without compromising the robustness and privacy of the data involved.
Our empirical evaluations substantiate that the proposed system not only sustains a high level of accuracy, comparable to models trained through conventional methods, but also improves or matches the robustness of adversarial training. The integration of distributed noise is pivotal, facilitating the safeguarding of each client's dataset while conforming to the standards of differential privacy.

The promising results obtained underscore the efficacy of our system in counteracting the threats from malicious entities in a federated setting, thus promoting a secure and dependable federated learning ecosystem. Given its enhanced performance and security attributes, our approach emerges as a practical solution for various applications, especially in the medical sector where the sanctity of data and privacy are of utmost importance.

We envision this work as a foundational step for future research endeavors exploring the realms of robustness and privacy in federated learning systems. It is our hope that the advancements made in this study will contribute significantly to the development and deployment of privacy-preserving, and robust federated learning models in real-world applications, paving the way for innovations in medical and other critical domains.

\section*{Acknowledgement}
This research is supported by KWF Kankerbestrijding and the Netherlands
Organisation for Scientific Research (NWO)  Domain AES, as part of their joint strategic research programme: Technology for Oncology IL. The collaboration project is co-funded by the PPP allowance made available by Health Holland, Top Sector Life Sciences \& Health, to stimulate public-private partnerships. 
\bibliographystyle{IEEEtran}
\bibliography{IEEEabrv1}

\end{document}